\documentclass[letterpaper,10 pt,conference]{ieeeconf}
\IEEEoverridecommandlockouts
\usepackage{algorithm}
\usepackage{algpseudocode}
\usepackage{amsmath}  
\usepackage{amssymb}  
\usepackage{bm}
\usepackage{epsfig}   
\usepackage{enumerate}
\usepackage{graphics} 
\usepackage{graphicx}
\usepackage{mathptmx} 
\usepackage{subfig}
\usepackage{url}

\graphicspath{{images/}}

\makeatletter
\renewcommand{\fnum@figure}{Fig. \thefigure}
\makeatother

\graphicspath{{images/}}

\title{\LARGE \textbf{Few-Shot Fruit Segmentation via Transfer Learning}}

\author{Jordan A.~James$^{1}$, Heather K.~Manching$^{2}$, Amanda M.~Hulse-Kemp$^{2,3}$, and William J.~Beksi$^{1}$%
\thanks{$^{1}$The authors are with the Department of Computer Science and Engineering,
The University of Texas at Arlington, Arlington, TX, USA.
Emails:
jaj9608@mavs.uta.edu,
william.beksi@uta.edu.
}
\thanks{$^{2}$The author is with the Department of Crop and Soil Sciences,
North Carolina State University, Raleigh, NC, USA.
Email:
hkmanchi@ncsu.edu.
}
\thanks{$^{3}$The author is with the Genomics and Bioinformatics Research Unit,
USDA Agricultural Research Service, Raleigh, NC, USA.
Email:
amanda.hulse-kemp@usda.gov.
}
}

\begin{document}

\maketitle
\pagestyle{empty}

\begin{abstract}
Advancements in machine learning, computer vision, and robotics have paved the
way for transformative solutions in various domains, particularly in
agriculture. For example, accurate identification and segmentation of fruits
from field images plays a crucial role in automating jobs such as harvesting,
disease detection, and yield estimation. However, achieving robust and precise
infield fruit segmentation remains a challenging task since large amounts of
labeled data are required to handle variations in fruit size, shape, color, and
occlusion. In this paper, we develop a few-shot semantic segmentation framework
for infield fruits using transfer learning. Concretely, our work is aimed at
addressing agricultural domains that lack publicly available labeled data.
Motivated by similar success in urban scene parsing, we propose specialized
pre-training using a public benchmark dataset for fruit transfer learning. By
leveraging pre-trained neural networks, accurate semantic segmentation of fruit
in the field is achieved with only a few labeled images. Furthermore, we show
that models with pre-training learn to distinguish between fruit still on the
trees and fruit that have fallen on the ground, and they can effectively
transfer the knowledge to the target fruit dataset. 
\end{abstract}


\section{Introduction}
\label{sec:introduction}
Automation, principally in the realm of infield fruit segmentation, has ushered
in a new era of modern agriculture. The integration of advanced technologies
such as  machine learning, computer vision, and robotics has started to
revolutionize the way we cultivate, harvest, and process fruits. Through
automated systems capable of accurately identifying and categorizing fruits
while still on the plant, efficiency has surged resulting in reduced costs and
minimized waste \cite{zhang2021orchard}. Nonetheless, despite these significant
strides, the task of infield fruit segmentation still remains a complex
challenge. The immense diversity of fruit shapes, sizes, colors, and textures
poses difficulties in developing universally applicable algorithms. Moreover,
the scarcity of labeled training data hinders the training of machine learning
models, as collecting and annotating large datasets for every fruit variety is a
resource-intensive endeavor.

\begin{figure}[t]
\centering
\subfloat[Input image.]{\includegraphics[scale=0.2]{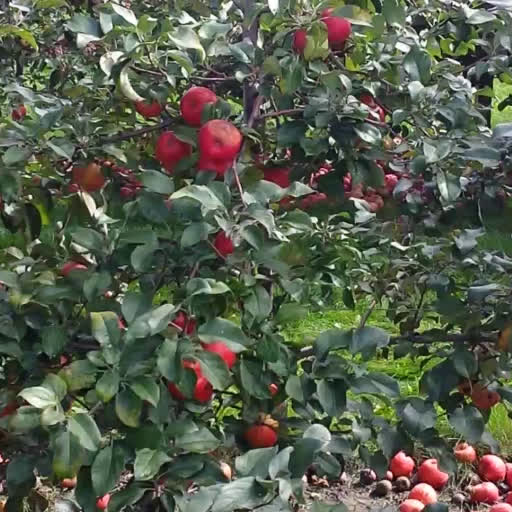}}
\hspace{1mm}
\subfloat[Ground-truth image.]{\includegraphics[scale=0.2]{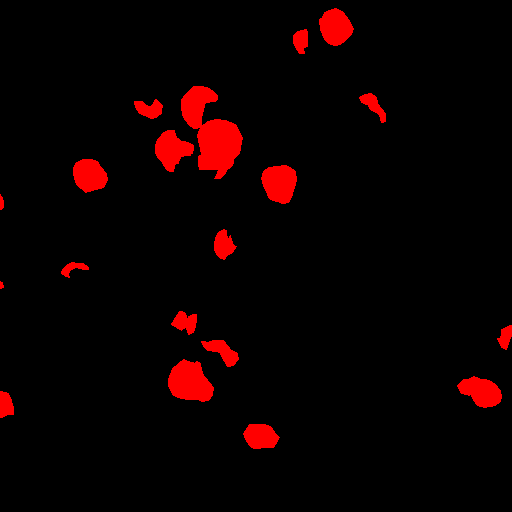}}
\hspace{1mm}
\subfloat[Two-shot CitDet results.]{\includegraphics[scale=0.2]{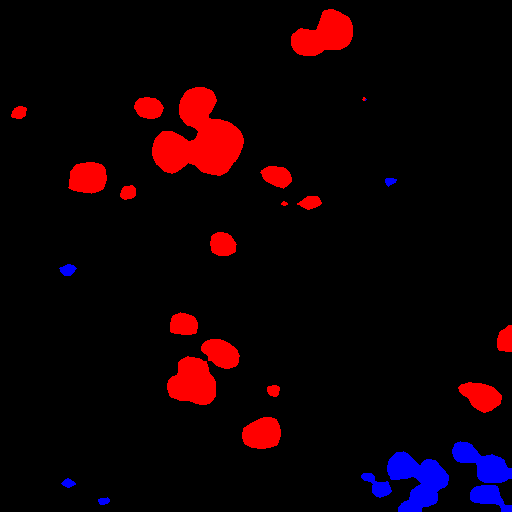}}
\hspace{1mm}
\subfloat[Two-shot ImageNet results.]{\includegraphics[scale=0.2]{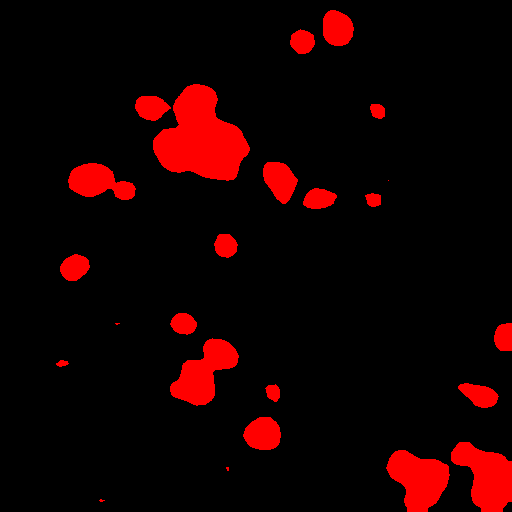}}
\caption{Qualitative results of our two-shot apple segmentation with specialized
pre-training on CitDet \cite{james2024citdet} and generalized pre-training on
ImageNet \cite{deng2009imagenet}. Predicted fruit on the tree are colored red
while predicted fruit on the ground are colored blue. Models pre-trained on
CitDet are capable of distinguishing between fruit on the ground and fruit on
the tree using only small amounts of labeled data to learn from.}
\label{fig:concept_overview}
\end{figure}

Few-shot learning (FSL) has emerged as a contemporary research area that allows
models to learn novel tasks from a few samples
\cite{parnami2022learning,catalano2023few}. In traditional machine learning,
models require substantial amounts of data to achieve satisfactory performance.
Conversely, FSL aims to enable models to generalize effectively even when
presented with just a handful of examples. Transfer learning, a technique
commonly employed in FSL, plays a pivotal role in this process. It involves
training a model on a source task with ample data and then transferring the
learned knowledge to a target task with limited data. This enables the model to
leverage its understanding of the source task to bootstrap its performance on
the target task. By pre-training on a large dataset and fine-tuning on the task
of interest, transfer learning minimizes the need for vast amounts of labeled
data. It accelerates the learning process for unique tasks, thus making it a
crucial approach in advancing the capabilities of machine learning models in
scenarios with limited data availability.

In this work, we show that accurate semantic segmentation of in-orchard apples
can be achieved with scarce amounts of data using FSL with transfer learning,
Fig.~\ref{fig:concept_overview}. Our method leverages a three-branch decoder for
direct learning of fruit boundaries and pre-training on CitDet
\cite{james2024citdet}, an open-source annotated image dataset of infield citrus
fruit. The network is evaluated on the MinneApple \cite{hani2020minneapple} test
set, which consists of 331 labeled images of in-orchard apples. In summary, our
contributions are the following.
\begin{itemize}
  \item A light-weight three-branch decoder that effectively learns task
  specific boundaries enabling knowledge transfer of learned shapes and
  boundaries between tasks.
  \item A framework for few-shot fruit segmentation that leverages specialized
  pre-training of a neural network and is designed to learn known features with
  similarities between fruits of different species such as the shape.
\end{itemize}
The source code associated with this project is publicly available at
\cite{fsfs}.

The remainder of the paper is organized as follows. We provide an overview of
related research in Section~\ref{sec:related_work}. The details of our framework
are presented in Section~\ref{sec:method}. Our evaluation results are discussed
in Section~\ref{sec:evaluation}. In
Section~\ref{sec:conclusion_and_future_work}, we conclude and provide directions
for future work.

\section{Related Work}
\label{sec:related_work}
Representative work towards FSL, fruit detection and segmentation, and real-time
semantic segmentation are discussed separately in this section.

\subsection{Few-Shot Learning} 
ImageNet \cite{deng2009imagenet}, an important benchmark dataset for computer
vision, contains a vast collection of labeled images across numerous categories.
Pre-training networks on ImageNet has proven to be an effective way of
initializing weights for a target task with a small sample size
\cite{sener2017active}. Multiple works have shown that classifiers with a higher
ImageNet accuracy achieve greater overall object detection and transfer accuracy
\cite{huang2017speed,kornblith2019better}. Furthermore, Hendrycks et al.
\cite{hendrycks2019using} showed that pre-training improves model robustness and
uncertainty. Huh et al. \cite{huh2016makes} also presented the advantages of
ImageNet pre-training even with the removal of classes related to the target
task. 

Although pre-training on ImageNet has many benefits, pre-training on specialized
datasets with high similarity to the target dataset can also improve accuracy.
One notable domain where this has been used to achieve state-of-the-art results
is urban scene parsing, which involves pre-training on the large Mapillary
Vistas \cite{neuhold2017mapillary} dataset and fine-tuning on Cityscapes
\cite{cordts2016cityscapes}. Specialized pre-training has been shown to benefit
the accuracy of semantic segmentation for various neural network architectures
\cite{xie2021segformer,li2019global,zhang2021dcnas}. In our work we show that
specialized pre-training also benefits the task of infield fruit segmentation.

\subsection{Fruit Detection and Segmentation}
Infield detection and segmentation of fruit is vital for agricultural tasks
including fruit yield, yield loss estimation, and automated picking. However,
variable lighting conditions, occlusions, and clustering make the accurate
recognition of fruit in an orchard environment very difficult. Over the last two
decades, many machine vision systems have been developed to address these
problems. Early techniques (e.g.,
\cite{zhao2005tree,hannan2009machine,arivazhagan2010fruit}) for automating fruit
recognition relied on hand-crafted features for specific fruits in order to
exploit unique attributes such as color, texture, and shape to separate fruit
from background foliage. 

More recently, methods utilizing deep learning have shown great promise in the
domain of fruit segmentation. For example, a fully convolutional neural network
\cite{long2015fully} was utilized by Bargoti and Underwood
\cite{bargoti2017image}, followed by watershed segmentation and a circular Hough
transform, to detect in-orchard apple instances from segmentation masks for
yield estimation. Semantic segmentation was employed by Chen et al.
\cite{chen2017counting} to detect and segment fruit blobs, which were then used
as input to a convolutional neural network for yield estimation. In addition,
Liu et al. \cite{liu2018robust} utilized a fully convolutional neural network to
segment and detect fruits in a tracking pipeline. 

The cost-efficient generation of synthetic data is also playing an increasing
important role in deep learning for agriculture \cite{kamilaris2018deep}. For
instance, Rahnemoonfar and Sheppard \cite{rahnemoonfar2017deep} demonstrated
that synthetic data can be used to train deep neural networks for counting
tomatoes in the field. In addition, Barth et al. \cite{barth2018data} used
synthetically generated data to train a DeepLab \cite{chen2017rethinking} model
to segment pepper images. Our approach distinguishes itself from prior research
efforts by harnessing acquired knowledge in fruit image segmentation,
specifically by employing knowledge gained from segmenting one fruit to enhance
the segmentation of another distinct fruit.

\subsection{Real-Time Semantic Segmentation}
Real-time semantic segmentation algorithms are necessary for practical
applications in agricultural automation that require fast interactions and
responses. Many methods adopt an encoder-decoder architecture such as SwiftNet
\cite{orsic2019defense} and DFANet \cite{li2019dfanet}. Notably, SwiftNet
provides information to a light-weight decoder from two separate branches each
processing feature maps at different resolutions. ShuffleSeg
\cite{gamal2018shuffleseg} proposes a decoder for ShuffleNet
\cite{zhang2018shufflenet} with skip connections and $1 \times 1$ convolutions
for channel transformations. The DeepLab \cite{chen2017rethinking} architecture
proposes atrous convolutions, an atrous spatial pyramid pooling module for the
decoder, and it adopts a high-resolution skip connection. FANet
\cite{hu2020real} modifies the traditional U-shape architecture with fast
attention in between the skip links. It uses a light-weight ResNet-18
\cite{he2016deep} backbone to attain real-time performance. 

Recent works have achieved real-time performance using parallel branched
networks. For instance, BiSeNetV1 \cite{yu2018bisenet} and BiSeNetV2
\cite{yu2021bisenet} employ a two-branch architecture consisting of a contextual
detail branch and a spatial detail branch. This architecture was further
enhanced by adding connections between branches in DDRNet \cite{hong2021deep},
and a deep aggregation pyramid pooling module (DAPPM) was added to efficiently
aggregate contextual details from the context branch. PIDNet \cite{xu2023pidnet}
proposed a three-branch network, which utilizes an auxiliary derivative branch
(ADB) to detect boundaries and guide the fusion between two other branches using
a boundary attention guided (BAG) fusion module. Further experiments showed the
benefit of adding the ADB-BAG method to previously proposed two-branch
architectures. Finally, a parallel aggregation pyramid pooling module (PAPPM)
inspired by the DAPPM was proposed for fast aggregation of contexts.

\section{Few-Shot Fruit Segmentation}
\label{sec:method}
\begin{figure*}[t]
\centering
\includegraphics[width=\textwidth,height=10.5cm]{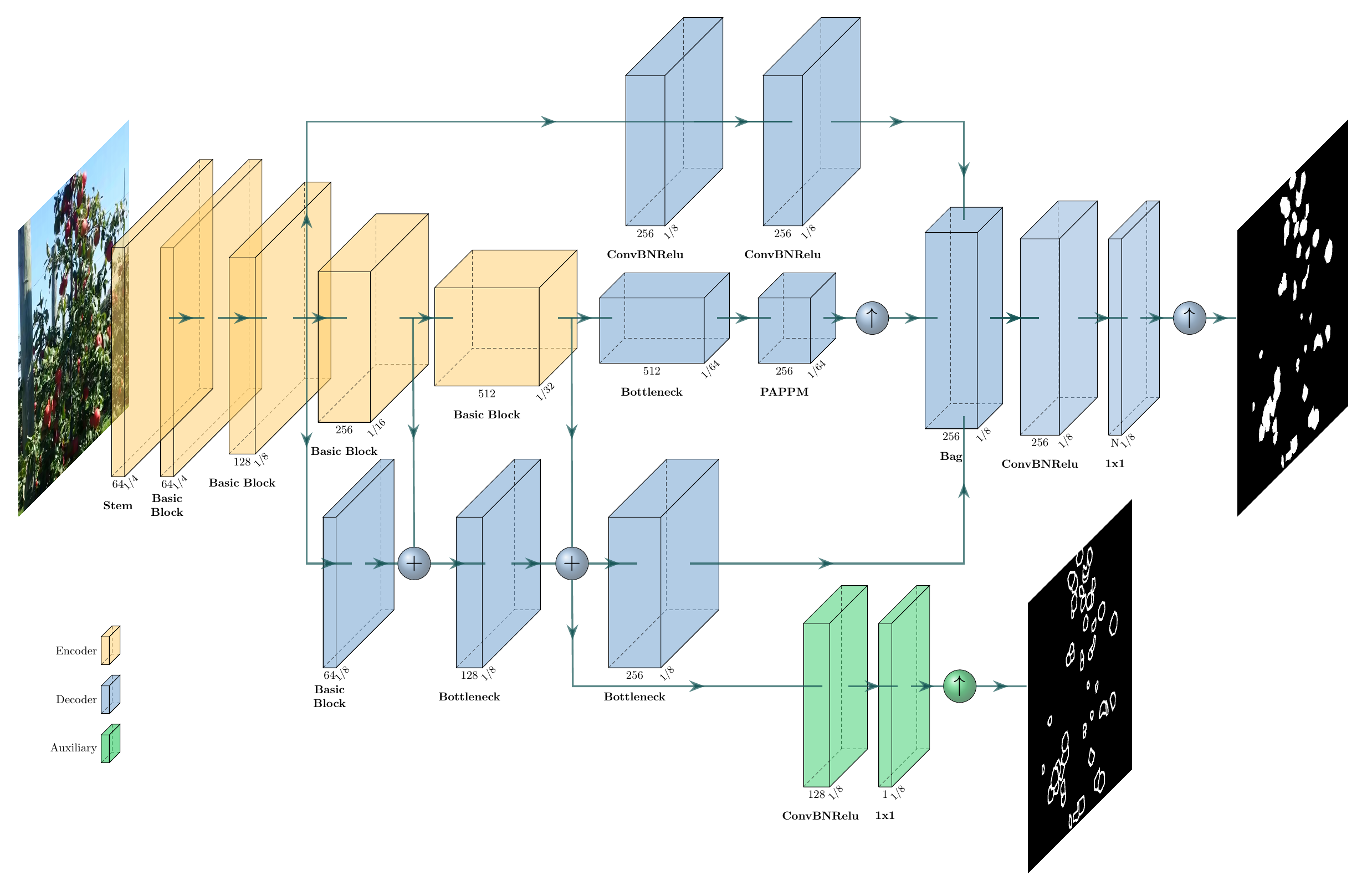}
\caption{The architecture of our few-shot fruit segmentation network. The
encoder, ResNet-18, is colored yellow, the decoder is blue, and the auxiliary
boundary head (train only) is green. The ``Basic Block" is the basic residual
block in ResNet-18, and the ``Bottleneck" is the bottleneck block from
ResNet-50. The ConvBNRelu blocks use 3 $\times$ 3 kernels for the convolutional
layers.}
\label{fig:architecture}
\end{figure*}

Our proposed framework is designed for fruit segmentation tasks with a scarce
amount of available labeled data. We take a simple approach, consisting of
pre-training on a similar task and standard data augmentation, to achieve
accurate few-shot segmentation. To further facilitate the task transfer by
learning the shape of the fruit, we use a three-branch decoder inspired by the
design of PIDNet \cite{xu2023pidnet}. ResNet-18 \cite{he2016deep} is adopted as
the backbone encoder for a light-weight design with 23 million trainable
parameters. The network architecture is depicted in Fig.~\ref{fig:architecture}.

\subsection{Three-Branch Decoder}
The decoder consists of a high-resolution spatial branch, a low-resolution
context branch, and an ADB. The feature maps are fused using a BAG fusion module
and then passed to a segmentation head to obtain the output mask.

\subsubsection{Spatial Branch}
For a light-weight design, the spatial branch relies heavily on the spatial
representations learned by the backbone network. The channels of the feature map
of the backbone with an output stride of 8 are transformed to match the channels
of the ADB and context branch with just two convolutional layers.

\subsubsection{Context Branch}
The final feature map of the backbone is used as input for the context branch.
Following the success of PIDNet \cite{xu2023pidnet} and DDRNet
\cite{hong2021deep}, we use a PAPPM at an output stride of 64 for efficient
aggregation of contexts. A bottleneck block is included before the PAPPM to
downsample the input feature map to the output stride of 64 for improved
computational efficiency. 

\subsubsection{ADB-BAG}
The ADB and BAG module are incorporated to enhance the knowledge transfer
between datasets by learning shapes and boundaries that are common between them.
To learn boundaries, the second stage of the ADB is inputted to an auxiliary
head and trained with binary cross-entropy loss (also denoted as the boundary
loss) on generated boundary labels. The boundary masks are obtained by using the
Canny edge detector \cite{canny1986computational} on the instance masks followed
by dilation. The BAG module then fuses the contextual and spatial information
along the learned boundaries by utilizing a sigmoid attention function.

\subsection{Knowledge Transfer from Specialized Tasks}
When dealing with specialized tasks or domains where collecting a large amount
of labeled data is challenging, transfer learning can be highly advantageous.
Instead of starting with a generic pre-trained model, the model is initialized
with weights from a network that was pre-trained on a dataset similar to the
target task. In this work, semantic segmentation of infield citrus is
transferred to the task of few-shot semantic segmentation of in-orchard apples.

\subsubsection{Datasets}
The CitDet \cite{james2024citdet} dataset is used for pre-training the model for
the task of citrus segmentation. CitDet contains many high-resolution images of
citrus trees with bounding box annotations for both fruit on the ground and on
the trees. To obtain masks for pre-training a semantic segmentation model, we
make use of Meta's Segment Anything Model \cite{kirillov2023segment} along with
the prompt encoder to obtain high-quality instance masks for objects within the
specified labels. The obtained instance masks are then combined and assigned the
class label according to the original bounding box annotation. The model is then
trained on images from CitDet resized to $1024 \times 1024$ with the obtained
masks and the best model weights, determined by the highest validation mean
intersection over union (mIoU), are saved for transfer learning. 

MinneApple \cite{hani2020minneapple} is chosen as the target dataset for
semantic segmentation of in-orchard apples. The dataset shares similarities with
CitDet. For instance, both datasets contain images of entire trees and labels of
fruit under varying amounts of occlusion. Yet, a key difference between the two
datasets is that the fruit on the ground class is not included in MinneApple.
Although the fruit on the ground class has no labels in the target dataset, we
include the class as an output to allow a direct transfer of the previously
learned distinctions between fruit on the ground and fruit on the tree.

\subsubsection{Data Augmentation}
\begin{figure}
\centering
\subfloat[Input image.]{\includegraphics[scale=0.11]{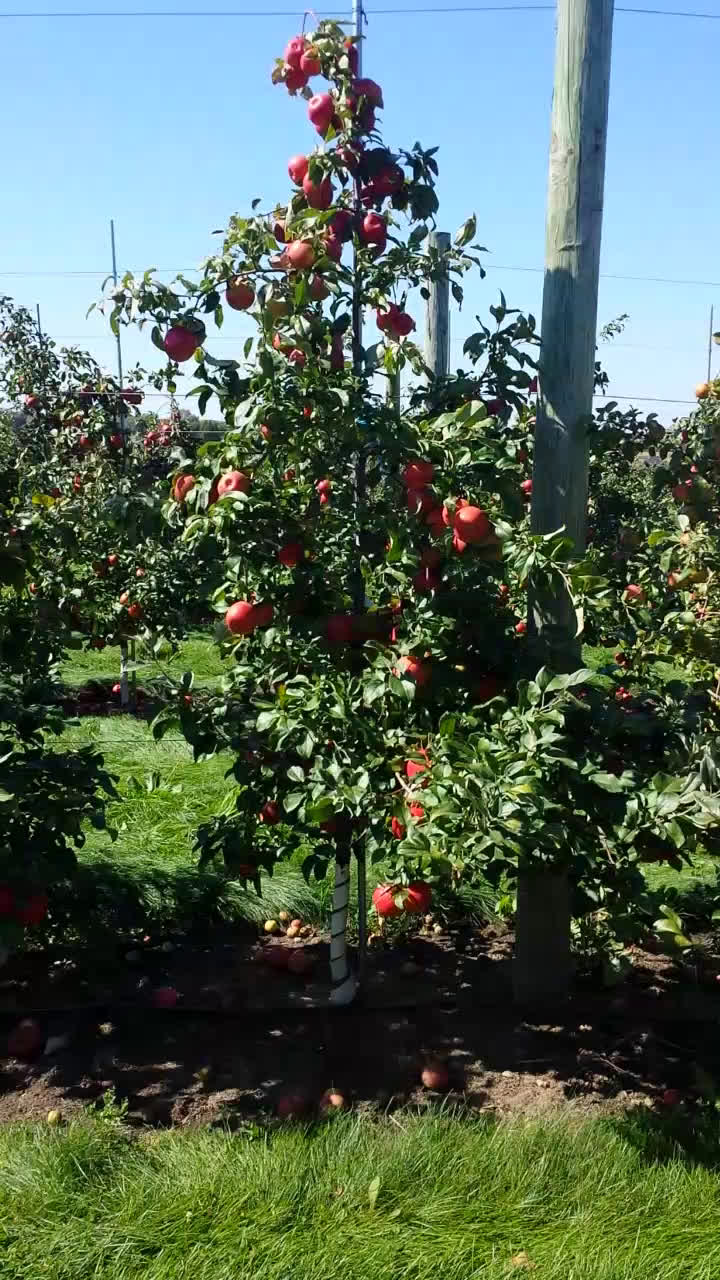}}
\hspace{1mm}
\subfloat[Augmented input images.]{\includegraphics[scale=0.088]{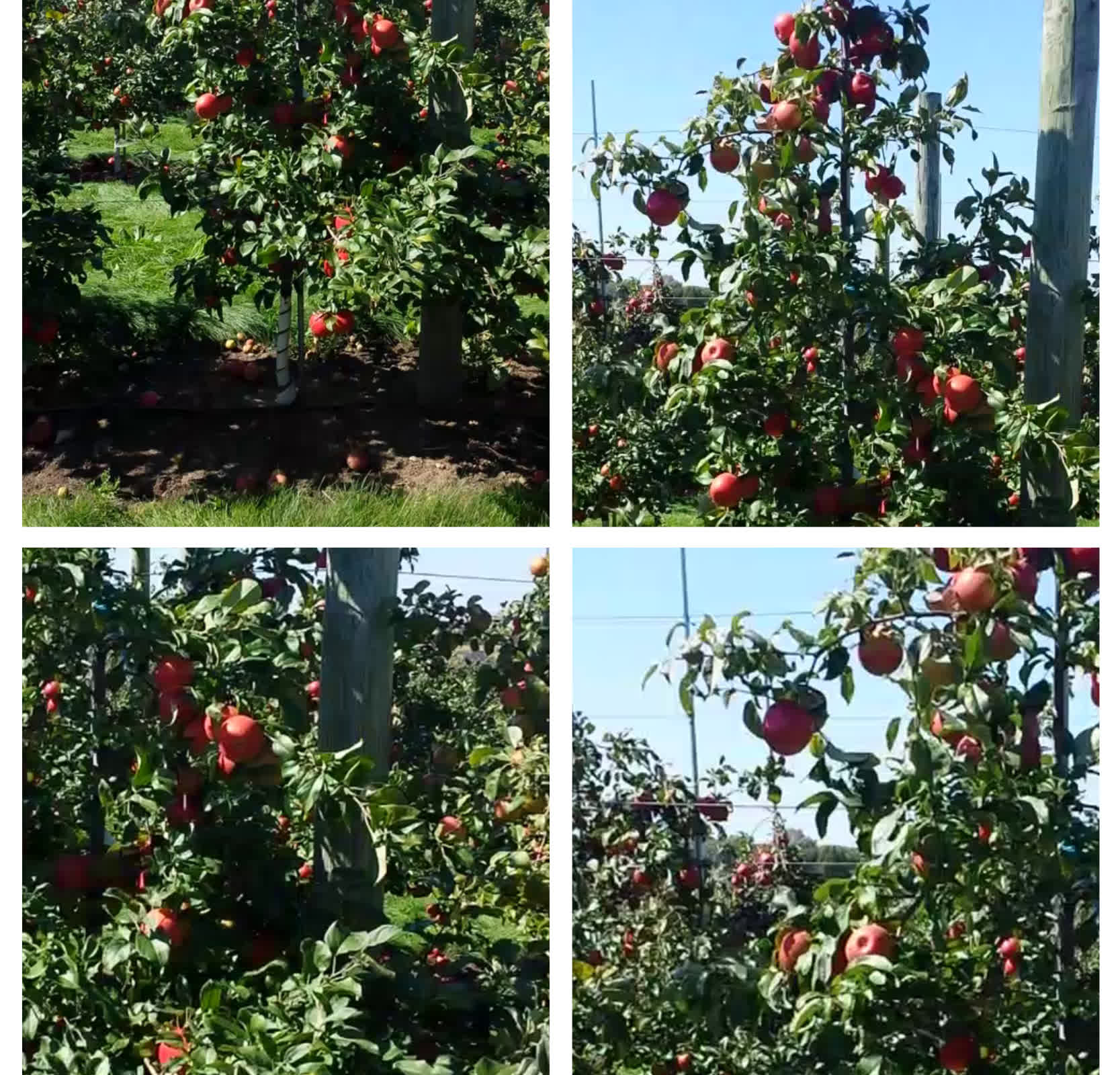}}
\caption{A visualization of the data augmentation used for few-shot semantic
segmentation of in-orchard apples. The original image is shown in (a) and the
resulting augmented image crops are shown in (b). Each panel in (b) displays a
random crop of the scale used where the top left is scaled by 0.75, top right by
1.0, bottom left by 1.25, and bottom right by 1.5.}
\label{fig:augmentation}
\end{figure}

To enable training with common batch sizes using sparse amounts of labeled data,
we apply data augmentation techniques before and during training. Before
training each image, a mask is scaled by [0.75, 1.0, 1.25, 1.5] and 5 random
crops of size $512 \times 512$ are taken from each scale to produce 20 training
images from each original image. Example visualizations of the random crops at
each scale can be seen in Fig.~\ref{fig:augmentation}. At training time, random
horizontal flipping is applied to prevent the model from learning to predict
fruit on only one side of an image.

\section{Evaluation}
\label{sec:evaluation}

\begin{figure*}
\centering
\subfloat[Input image.]{\includegraphics[scale=0.20]{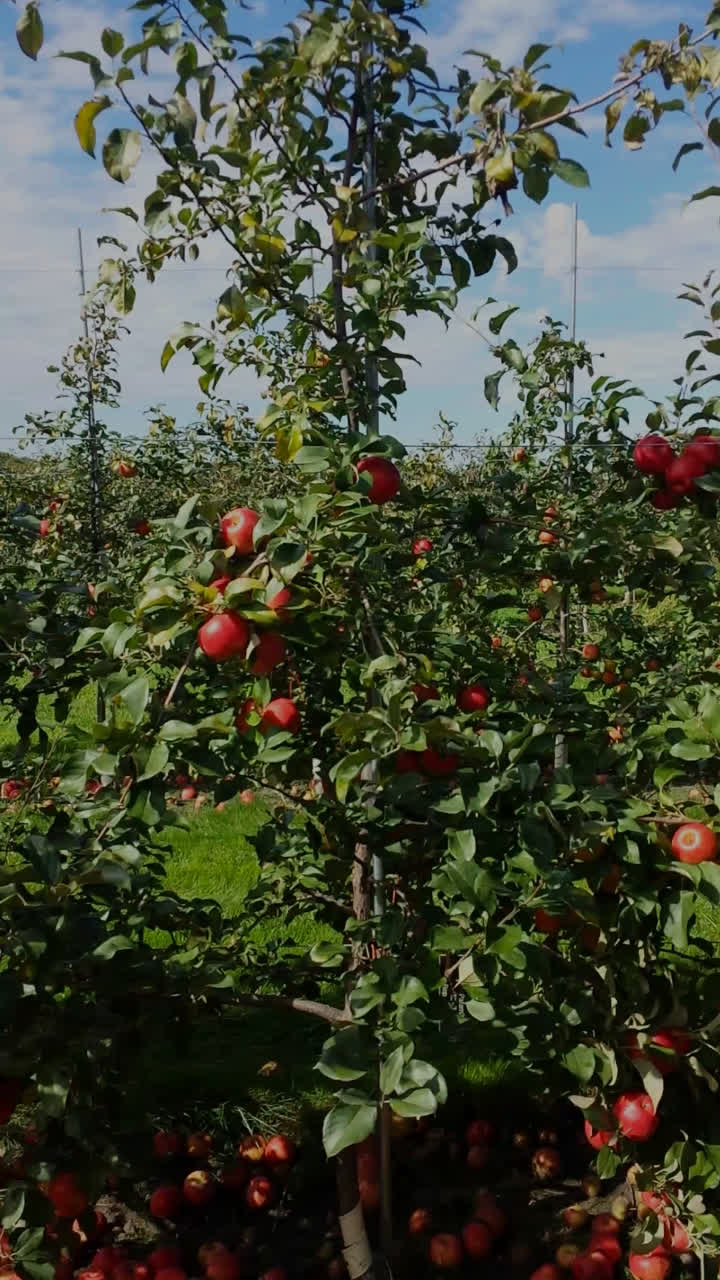}}
\hspace{1mm}
\subfloat[Ground-truth image.]{\includegraphics[scale=0.20]{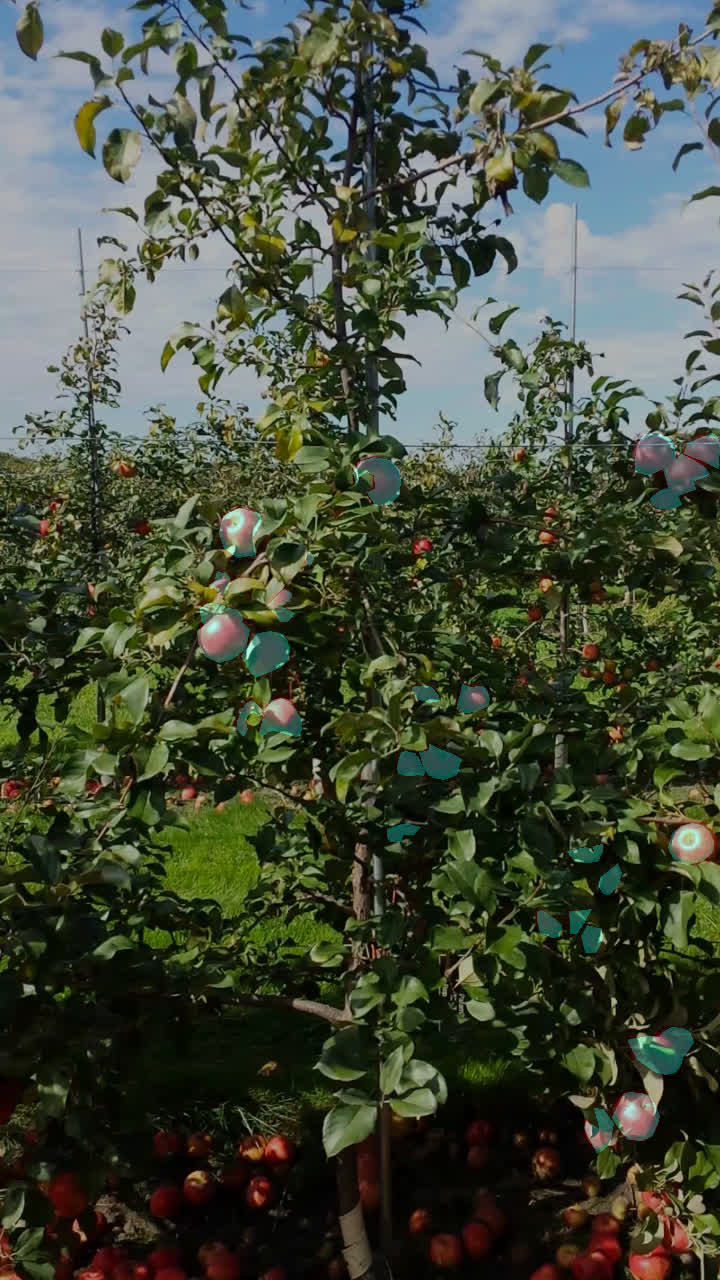}}
\hspace{1mm}
\subfloat[Prediction images.]{\includegraphics[scale=0.085375]{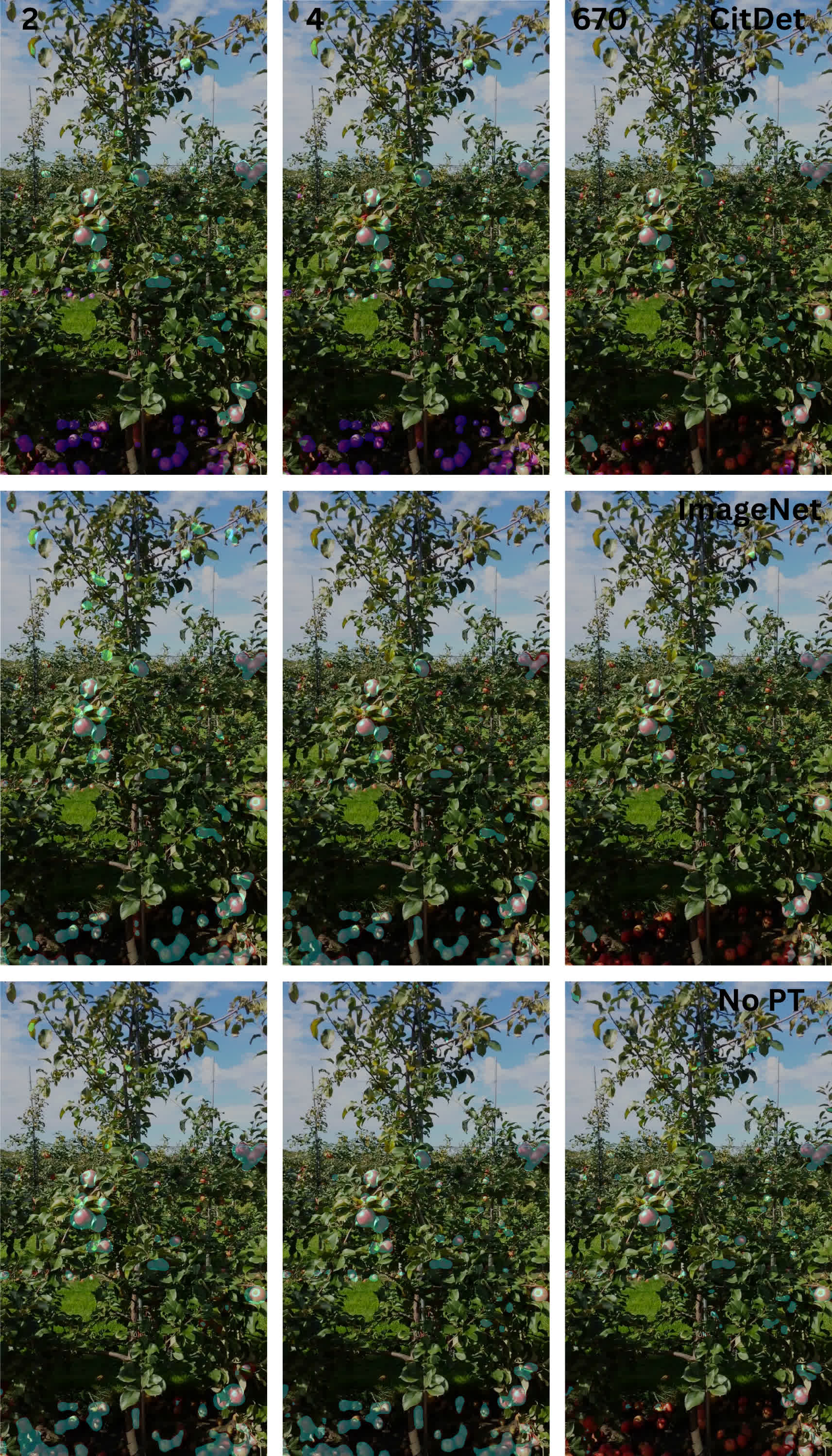}}
\caption{Qualitative results for few-shot and full-shot apple segmentation for
each pre-training method. Predicted fruit on the tree are colored cyan and
predicted fruit on the ground are colored magenta. The original input can be
seen in (a), the ground-truth labels in (b), and predictions in (c). Each row of
predictions correspond to the pre-training method. The top row is CitDet, the
middle row is ImageNet, and the bottom row is no pre-training. Each column
corresponds to the number of training images used for fine tuning (i.e., 2, 4,
and 670), increasing from left to right (best viewed zoomed in).}
\label{fig:qualitative_results}
\end{figure*}

\begin{figure}
\centering
\includegraphics[width=\columnwidth]{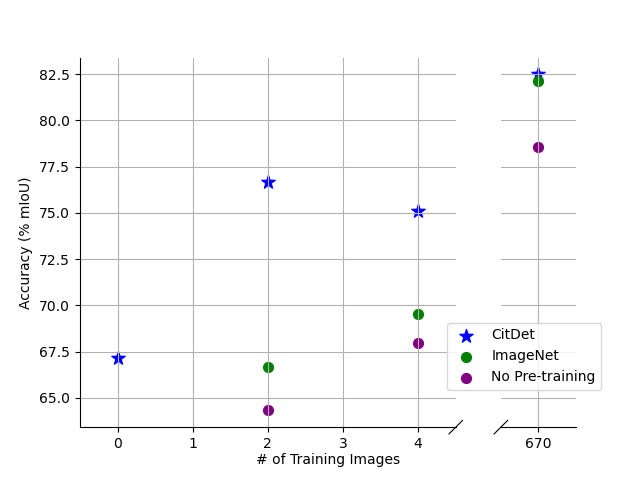}
\caption{The mIoU performance on the MinneApple test set versus the number of
training images used from the MinneApple training set. Standard pre-training
methods are marked as dots, and our specialized pre-training is shown in blue
stars. Our method achieves a better accuracy than traditional approaches,
especially with only a few annotated images.}
\label{fig:eval_plot}
\vspace{-2mm}
\end{figure}

\begin{figure}
\centering
\subfloat[Zero-shot results.]{\includegraphics[scale=0.1575]{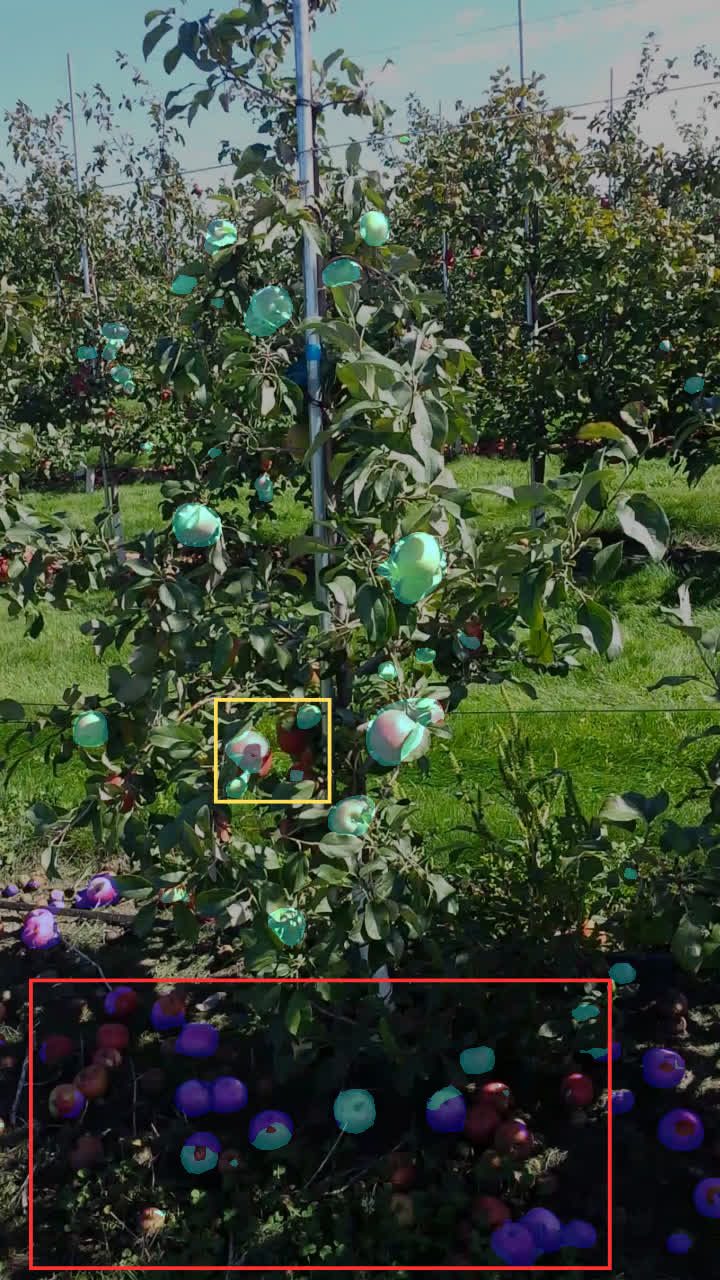}}
\hspace{1mm}
\subfloat[Two-shot results.]{\includegraphics[scale=0.1575]{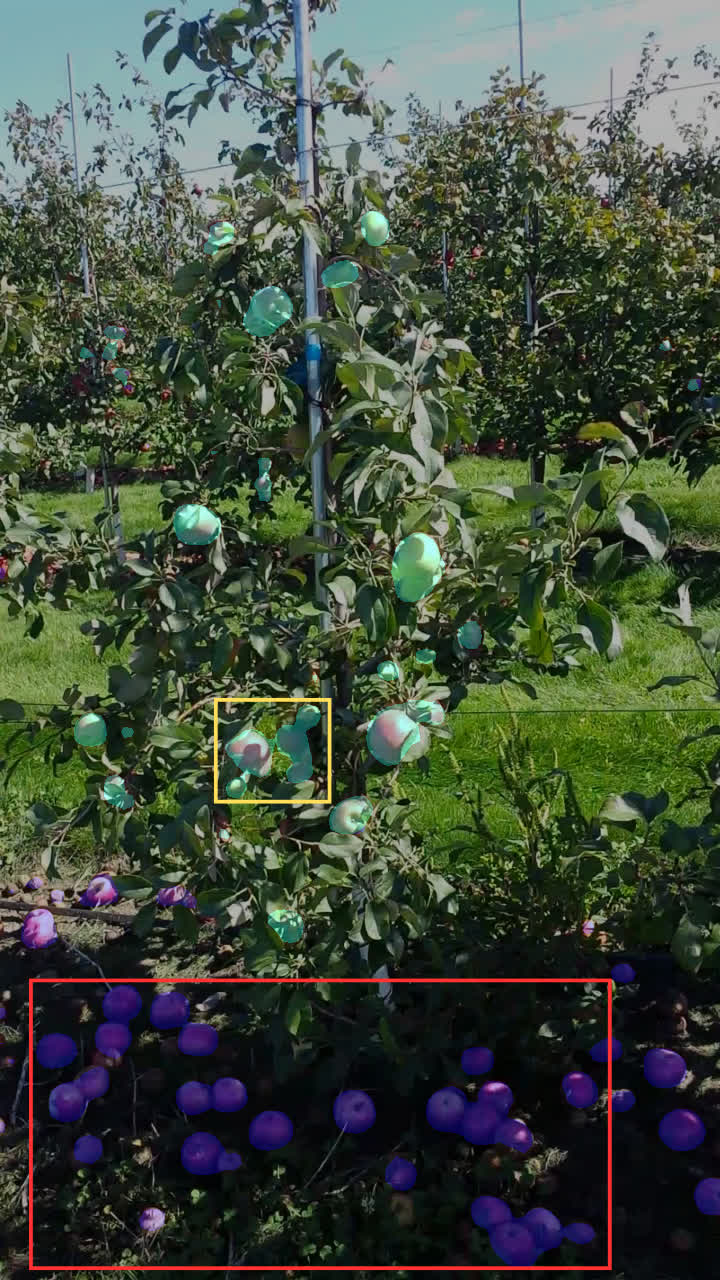}}
\caption{Qualitative results for zero-shot and two-shot apple segmentation.
Predicted fruit on the tree are colored cyan and predicted fruit on the ground
are colored magenta. The red box highlights the unsupervised learning of the
fruit on the ground class. With only two images, highly-occluded fruit are
detected as shown in the yellow boxes.}
\label{fig:few_shot_vis}
\vspace{-2mm}
\end{figure}

\begin{table}
\centering
\caption{Quantitative results on the MinneApple test set with CitDet
pre-training.}
\resizebox{7.25cm}{!}{\begin{tabular}{cccc}
Pre-training & \# of Training Images & mIoU & PA \\ 
\cline{1-4}
CitDet     & 0   & 67.2 & 97.3 \\
CitDet     & 2   & 76.7 & 98.0 \\
CitDet     & 4   & 75.1 & 98.1 \\
CitDet     & 670 & 82.5 & 98.7
\end{tabular}}
\label{tab:table_1}
\end{table}

\begin{table}
\centering   
\caption{Quantitative results on the MinneApple test set with ImageNet
pre-training.}
\resizebox{7.25cm}{!}{\begin{tabular}{cccc}
Pre-training & \# of Training Images & mIoU & PA  \\ 
\cline{1-4}
ImageNet & 2   & 66.7 & 95.8 \\
ImageNet & 4   & 69.5 & 96.9 \\
ImageNet & 670 & 82.1 & 98.7 
\end{tabular}}
\label{tab:table_2}
\end{table}

\begin{table}
\centering
\caption{Quantitative results on the MinneApple test set with no pre-training.}
\resizebox{7.25cm}{!}{\begin{tabular}{cccc}
Pre-training & \# of Training Images & mIoU & PA  \\ 
\cline{1-4}
None & 2   & 64.3 & 96.1 \\
None & 4   & 67.9 & 96.8 \\
None & 670 & 78.6 & 98.3 
\end{tabular}}
\label{tab:table_3}
\end{table}

In this section, we experimentally assess our method on few-shot semantic
segmentation of in-orchard apples using two and four training images, and
evaluate on zero-shot transfer using CitDet pre-training. The class-wise mIoU
and the percent of pixels that are accurately classified in the image, i.e.,
pixel accuracy (PA), are chosen as the evaluation metrics. All experiments were
evaluated on the MinneApple test set containing 331 images. For all models, the
fruit on the ground class was set to the background before evaluation. 

The quantitative results on the MinneApple test set are shown in the
Table~\ref{tab:table_1}. We describe the implementation details and compare our
results with typical full-shot semantic segmentation in
Section~\ref{subsec:implementation_details}. Additionally, we conduct
experiments using traditional ImageNet pre-training as well as no pre-training
with randomly initialized weights and report the results in
Table~\ref{tab:table_2} and Table~\ref{tab:table_3}, respectively.
Section~\ref{subsec:ablation_study} provides an ablation study. The evaluation
results and key findings are discussed in Section~\ref{subsec:discussion}. 

\subsection{Implementation Details}
\label{subsec:implementation_details}
Our implementation was developed using PyTorch and the Lightning
\cite{Falcon_PyTorch_Lightning_2019} package. We utilized the timm open-source
library to import the pre-trained weights from the ImageNet dataset for the
ResNet encoder. For all methods and datasets, a ``poly" learning rate scheduled
with a factor of 0.9 was used. The stochastic gradient descent optimizer with a
weight decay of 1E-4 and momentum of 0.9 was employed with a batch size of 4 for
fine-tuning. The proposed data augmentation was performed before training and
the same random crops were used for all experiments.

Hyperparameters, such as the number of epochs and starting learning rate, were
held constant between experiments with a varying number of training images. In
all experiments the model with the best performance on the test set was saved.
For models initialized with random weights and no pre-training, the starting
learning rate was set to 7.5E-3 and each model was trained for 100 epochs. For
experiments with ImageNet, the pre-trained encoder weights were imported from
timm and the entire network was fine-tuned at a starting learning rate of 1E-3
for 50 epochs. Finally, transfer from CitDet was achieved by importing
pre-trained weights for both the encoder and decoder, and fine-tuned with a
starting learning rate of 1E-4 for 50 epochs.

\subsection{Ablation Study}
\label{subsec:ablation_study}
\begin{table}
\centering
\caption{Ablation study results for a two-branch architecture.}
\resizebox{7.25cm}{!}{\begin{tabular}{cccc}
Pre-training & \# of Training Images & mIoU & PA  \\ 
\cline{1-4}
CitDet    & 0   & 63.8 & 95.3 \\
ImageNet  & 2   & 60.4 & 92.7 \\
CitDet    & 2   & 71.6 & 95.7 \\
ImageNet  & 4   & 65.0 & 94.9 \\
CitDet    & 4   & 72.8 & 95.7 \\
ImageNet  & 670 & 73.4 & 96.3 \\
CitDet    & 670 & 73.9 & 95.9
\end{tabular}}
\label{table:ablation}
\end{table}
To validate the design of the three-branch decoder and separate the effects of
our specialized pre-training from the effects of the architecture, we performed
an ablation study on the ADB and corresponding BAG fusion module. Specifically,
we removed the ADB and replaced the BAG fusion module with the addition of the
spatial and context branches to create a two-branch decoder. The same training
and pre-training methods were adopted for the two-branch decoder. The models
were then evaluated on the MinneApple test set using the mIoU and PA metrics as
in the previous experiments. The results in Table~\ref{table:ablation}
demonstrate that pre-training on CitDet also greatly benefits few-shot semantic
segmentation of in-orchard apples for architectures without the ADB-BAG modules.
Comparing the quantitative results of the two-branch and three-branch
architectures, the inclusion of the ADB-BAG modules shows a large increase in
performance for both few-shot and full-shot segmentation. 

\subsection{Discussion}
\label{subsec:discussion}
We observe from the qualitative visualizations in
Fig.~\ref{fig:concept_overview} and Fig.~\ref{fig:qualitative_results} that
few-shot models without specialized pre-training were capable of segmenting
nearly all apples from the image. The main shortcoming of these models compared
to those pre-trained on CitDet is the inability to distinguish the difference
between labeled fruit on the trees and unlabeled fruit on the ground. This could
be a problem for applications requiring low latency such as fruit picking, as
post-processing with traditional computer vision methods would need to be used
to distinguish between the two classes. For all the pre-training methods, we
noticed that the boundaries of the masks became more refined as more samples
were added to the training set.

Similar to previous work, we found that pre-training using ImageNet
significantly improved the accuracy of semantic segmentation. The quantitative
results in Table~\ref{tab:table_2} and Table~\ref{tab:table_3} show that
pre-training benefits the task transfer for both few-shot and full-shot
segmentation. Additionally, we confirmed that specialized pre-training on
similar datasets greatly benefits few-shot fruit segmentation and it also
provides a slight increase in accuracy for full-shot segmentation as seen in
Table~\ref{tab:table_1} and Fig.~\ref{fig:eval_plot}. We can also see that
pre-training on CitDet allows for accurate zero-shot segmentation of in-orchard
apples and it even outperforms the two-shot model with a traditional ImageNet
transfer. We note that the two-shot CitDet model outperforms the four-shot
model. This indicates that the model can transfer to the task of fruit
segmentation with only a few examples, but it requires a larger number of
samples to learn the difference in annotation styles between the pre-training
and target datasets. The results of the ablation study in
Table~\ref{table:ablation} further shows that the addition of the ADB-BAG is
responsible for the FSL of the fruit segmentation task.

Interestingly, the predictions of the models pre-trained on CitDet for the fruit
on the ground class improved when fine-tuning on a small number of labeled
images despite only having labels for fruit on the trees. This demonstrates that
models can adapt the knowledge learned from a pre-training dataset in an
unsupervised manner to the target dataset. The unsupervised learning can be seen
in the qualitative results of Fig.~\ref{fig:few_shot_vis}. We hypothesize that
this unsupervised learning is able to occur due to the similarity between the
two fruit classes. As seen in the qualitative visualizations, models first learn
to predict all the fruit as fruit on the tree, indicating that the fruit on the
ground class is closely related. Since the CitDet models have already learned
the distinction between fruit on the tree and fruit on the ground, the
similarities between the two classes allow for unsupervised learning of one
without labels. This could suggest that the two classes fall under a
hierarchical fruit class in which all fruit is first identified before being
further classified by location.

\section{Conclusion and Future Work}
\label{sec:conclusion_and_future_work}
In this paper we proposed a framework for few-shot fruit segmentation consisting
of a three-branch decoder. The decoder was designed for detecting similar shape
between fruits and leverages specialized pre-training on a dataset with a high
similarity to the target. Our results indicate that specialized pre-training
significantly improves few-shot semantic segmentation for in-orchard apples and
gives a slight benefit over ImageNet pre-training for full-shot segmentation.
Since this specialized pre-training has seen benefits in other areas, future
few-shot segmentation work in all domains should consider the applicability of
specialized pre-training for the intended target task. The qualitative results
of unsupervised learning of the fruit on the ground class indicates potential
future work for unsupervised transfer learning between classes included in both
the pre-training and target datasets, but only labeled in the pre-training
dataset.

\section*{Acknowledgments} 
This material is based upon work supported by the United States Department of
Agriculture (USDA) under Agricultural Research Service CRIS Project
\#6066-21310-006-000-D and agreement \#58-6066-3-050.

\bibliographystyle{IEEEtran}
\bibliography{IEEEabrv,few-shot_fruit_segmentation_via_transfer_learning}

\end{document}